\documentclass[11pt]{article}

\usepackage[preprint]{acl}

\usepackage{times}
\usepackage{latexsym}

\usepackage[T1]{fontenc}

\usepackage[utf8]{inputenc}

\usepackage{microtype}

\usepackage{inconsolata}

\usepackage{graphicx}

\usepackage{booktabs}
\usepackage{multirow}
\usepackage{tabularx}
\usepackage{ragged2e}
\usepackage[table,dvipsnames]{xcolor}
\usepackage{threeparttable}

\usepackage{tikz}
\usetikzlibrary{positioning, arrows.meta, calc}

\usepackage{tcolorbox}
\tcbuselibrary{breakable}

\usepackage{amsmath}
\usepackage{amssymb}

\usepackage{enumitem}

\usepackage{url}
\usepackage{hyperref}
\hypersetup{hidelinks}

\usepackage{orcidlink}

\newcommand{\medrect}{\textsc{MedRECT}}
\newcommand{\medec}{\textsc{MEDEC}}
\newcommand{\medrectrepo}{\url{https://github.com/pfnet-research/medrect}}
\newcommand{\medrectzenodo}{\url{https://doi.org/10.5281/zenodo.17550933}}
\newcommand{\medrectmodel}{\url{https://huggingface.co/pfnet/Preferred-MedRECT-32B}}
\newcommand{\medecrepo}{\url{https://github.com/abachaa/MEDEC}}

\newcommand{\nummodels}{11}
\newcommand{\numconfigs}{17}

\title{\medrect{}: A Bilingual Medical Reasoning Benchmark \\ for Error Correction in Clinical Texts}

\author{Naoto Iwase~\orcidlink{0009-0002-7193-3468}\thanks{This work was done when N. Iwase worked at Preferred Networks, Inc. as a part-time engineer.} \\
  Nagoya University \\
  Preferred Networks, Inc. \\
  \texttt{naoto@iwase.dev} \\\And
  Hiroki Okuyama \\
  Preferred Networks, Inc. \\
  ~ \\
  \texttt{hokuyama@preferred.jp} \\\And
  Junichiro Iwasawa~\orcidlink{0000-0002-2560-5650} \\
  Preferred Networks, Inc. \\
  ~ \\
  \texttt{iwasawa@preferred.jp} \\}

\date{}

\begin{document}
\maketitle

\begin{abstract}
Large language models (LLMs) show promise in medical applications, but their ability to \emph{detect and correct errors in clinical texts} remains under-evaluated, particularly beyond English. We introduce \medrect{}, a bilingual benchmark for Japanese and English that formulates medical error handling as three subtasks: error detection, error sentence extraction, and error correction. \medrect{}-ja contains 663 samples derived from the Japanese Medical Licensing Examinations, while the separately sourced \medrect{}-en contains 458 samples curated from \medec{}. We evaluate \nummodels{} LLMs across \numconfigs{} configurations that cover proprietary and open-weight models, medical-domain specialization, and multiple reasoning settings. Qwen3-32B scores higher in its thinking mode than in its non-thinking mode on error detection F1 and sentence extraction accuracy in both subsets, with sentence extraction accuracy higher by 24.5 percentage points on \medrect{}-ja and 10.3 on \medrect{}-en. Several leading general-purpose reasoning models outperform all three evaluated medical-domain models on these two subtasks. Most models have lower point estimates on the Japanese subset, although absolute scores are not directly comparable because the subsets differ in source material and error distributions. LoRA fine-tuning yields higher sentence extraction accuracy and higher point estimates on all three reference-based correction similarity metrics in both languages. \medrect{} provides an open, reusable evaluation resource for studying medical error correction and reasoning across Japanese and English. Our dataset and code are available at \medrectrepo{}.
\end{abstract}

\section{Introduction}
\label{sec:intro}

\begin{figure*}[t]
\centering
\begin{tikzpicture}[
    node distance=0.4cm,
    textbox/.style={draw, rounded corners=3pt, text width=0.42\textwidth, align=left, font=\small, inner sep=6pt, fill=gray!5},
    taskbox/.style={draw, rounded corners=3pt, text width=0.28\textwidth, align=center, font=\small, inner sep=5pt},
    tasklabel/.style={font=\small\bfseries, inner sep=3pt},
    arrow/.style={-{Stealth[length=5pt]}, thick, gray!60},
    flowarrow/.style={-{Stealth[length=5pt]}, thick, black!50},
]

\node[textbox, text width=0.93\textwidth, fill=blue!3, line width=0.6pt] (clintext) {
    \textbf{Clinical Text (Input)}\\[3pt]
    \textsf{S1:} A 26-year-old male presents with concerns about his vision.\\
    \textsf{S2:} He reports difficulty judging how well meat is cooked and has chosen items he thought were light blue, only to be told they were pink.\\
    \textsf{S3:} \colorbox{red!25}{A \textbf{\color{red!70!black}binocular vision test} was performed as a diagnostic workup, which showed good stereopsis.}\\
    \textsf{S4:} His brother also reports similar difficulties with color perception.\\
    \textsf{S5:} Eye movement and pupillary responses were normal.
};

\node[taskbox, text width=0.22\textwidth, below left=0.8cm and 0.3cm of clintext.south, fill=orange!6, xshift=-3.0cm, line width=0.6pt] (task1) {
    \textbf{Does this text contain}\\
    \textbf{a medical error?}\\[6pt]
    {\Large \color{red!70!black} \textbf{Yes}}\\[-1pt]
    {\footnotesize (Error detected)}
};

\node[taskbox, below=0.8cm of clintext.south, fill=yellow!6, line width=0.6pt] (task2) {
    \textbf{Which sentence}\\
    \textbf{contains the error?}\\[8pt]
    {\Large \color{red!70!black} \textbf{S3}}\\[-1pt]
    {\footnotesize (Sentence 3 contains the error)}
};

\node[taskbox, below right=0.8cm and 0.3cm of clintext.south, fill=green!6, xshift=3.0cm, line width=0.6pt] (task3) {
    \textbf{How should it be corrected?}\\[8pt]
    {\small\color{ForestGreen} ``A \textbf{color vision test} was performed, confirming congenital red-green color vision deficiency.''}
};

\node[tasklabel, above=1pt of task1.north, fill=orange!25, rounded corners=2pt] {\scriptsize\textbf{Step 1: Error Detection}};
\node[tasklabel, above=1pt of task2.north, fill=yellow!30, rounded corners=2pt] {\scriptsize\textbf{Step 2: Error Sentence Extraction}};
\node[tasklabel, above=1pt of task3.north, fill=green!20, rounded corners=2pt] {\scriptsize\textbf{Step 3: Error Correction}};

\draw[arrow] (clintext.south -| task1.north) -- ([yshift=0.38cm]task1.north);

\draw[flowarrow] (task1.east |- task2.west) -- (task2.west) node[midway, above, font=\scriptsize\itshape] {if error};
\draw[flowarrow] (task2.east) -- (task3.west |- task2.east) node[midway, above, font=\scriptsize\itshape] {then};

\end{tikzpicture}
\caption{Overview of the \medrect{} benchmark, illustrated with a \medrect{}-ja sample (translated to English). Given a clinical text, models must (1)~detect whether it contains a medical error, (2)~identify the erroneous sentence, and (3)~generate a corrected version. Here, the patient's symptoms (difficulty distinguishing colors) and family history suggest congenital color vision deficiency, but a binocular vision test was ordered instead of the appropriate color vision test.}
\label{fig:task_overview}
\end{figure*}

Large language models (LLMs) offer unprecedented potential to augment clinical decision-making~\citep{Usuyama2025}, yet their deployment faces a critical challenge: the opacity and reliability of their reasoning processes. Models may arrive at correct conclusions through flawed logic~\citep{Turpin2023, lyu-etal-2023-faithful} or replicate cognitive biases (such as anchoring and confirmation bias) that contribute to diagnostic errors~\citep{Saposnik2016,Makary2016}.

While state-of-the-art LLMs demonstrate remarkable success on structured examinations like the USMLE~\citep{Gilson2023,Singhal2023}, multiple-choice question-answering (MCQA) performance insufficiently represents real-world clinical reasoning. Recent analyses highlight a critical gap between generating plausible text and ensuring patient safety~\citep{Moell2025}, underscoring the need for benchmarks that evaluate clinically relevant capabilities such as error detection and correction, not just question-answering accuracy.

This challenge is particularly acute in Japanese medical natural language processing (NLP). While Japanese medical LLMs advance rapidly~\citep{kawakami2025stabilizing,sukeda2024development}, the field has lacked standardized benchmarks for complex clinical tasks~\citep{jiang2024jmedbenchbenchmarkevaluatingjapanese}, and the English-only \medec{} benchmark~\citep{ben-abacha-etal-2024-overview,abacha2025medecbenchmarkmedicalerror} relies on resource-intensive manual annotation that does not scale to other languages.

To bridge these gaps, we introduce \medrect{} (A Bilingual \textbf{Med}ical \textbf{R}easoning Benchmark for \textbf{E}rror \textbf{C}orrection in Clinical \textbf{T}exts), to our knowledge the first benchmark for medical error detection and correction in Japanese, paired with English evaluation under a shared task definition (Figure~\ref{fig:task_overview}). Our contributions are: \textbf{(1) Open Benchmark Construction}: A reproducible pipeline that creates error correction data from the Japanese Medical Licensing Examinations (JMLE), accompanied by benchmark data and code. \textbf{(2) Bilingual Evaluation}: We evaluate \nummodels{} LLMs across \numconfigs{} configurations on \medrect{}-ja (663 samples) and \medrect{}-en (458 samples) under a shared task definition. \medrect{}-ja is derived from JMLE, whereas \medrect{}-en is drawn from the MedQA-based \medec{} MS subset of U.S. medical licensing exam questions. Neither subset is a translation of the other. \textbf{(3) Parameter-Efficient Fine-Tuning}: LoRA (Low-Rank Adaptation) fine-tuning~\citep{hu2021loralowrankadaptationlarge} improves sentence extraction accuracy and yields higher point estimates on all three reference-based correction similarity metrics in both languages.

\section{Related Work}

\subsection{Benchmarks for Medical Reasoning}

Medical LLM evaluation has centered on MCQA benchmarks from licensing exams, including MedQA~\citep{jin2020diseasedoespatienthave}, multilingual variants~\citep{Alonso2024}, and MultiMedQA with Med-PaLM achieving expert-level performance~\citep{Singhal2023}. However, MCQA cannot evaluate practical clinical tasks such as communication~\citep{Zeng2020MedDialog}, note generation~\citep{Tang2023Dialogue2Note,Yim2023ACIBench,VanVeen2024Summarization}, or error correction~\citep{abacha2025medecbenchmarkmedicalerror}. Beyond MCQA, clinical natural language inference performance varies across prompt structures, and LoRA effects differ by reasoning type~\citep{jullien-etal-2026-dissecting}.

\subsection{Error Detection and Correction in Clinical Texts}

The \medec{} benchmark~\citep{ben-abacha-etal-2024-overview,abacha2025medecbenchmarkmedicalerror} introduced systematic error correction evaluation with five clinical error types for the MEDIQA-CORR 2024 shared task. The closest multilingual benchmark is \textsc{MedErrBench}~\citep{ma2026mederrbench}, which contains 2,506 English, Chinese, and Arabic samples for the same three tasks, uses ten clinician-informed error types, and has all samples independently reviewed by two clinicians. \medrect{} complements these resources with Japanese coverage, a reusable automated construction pipeline based on JMLE, and analyses of reasoning settings and bilingual fine-tuning. Its English subset provides evaluation under the same task definition using quality-screened \medec{} data.

\subsection{Japanese Medical NLP and Multilingual Evaluation}

Japanese medical NLP resources include clinical BERT~\citep{Kawazoe2021}, MedWeb corpus~\citep{Wakamiya2019}, and JMedBench with 20 datasets including IgakuQA (JMLE 2018--2022) and primarily GPT-4 translated biomedical data~\citep{kasai2023evaluating, jiang2024jmedbenchbenchmarkevaluatingjapanese}. For multilingual evaluation, prior work explored medical QA performance variations across languages~\citep{Jin2023, Alonso2024} and biomedical entity linking~\citep{liu2021learningdomainspecialisedrepresentationscrosslingual}. To our knowledge, \medrect{} is the first benchmark for medical error detection and correction in Japanese and pairs this new coverage with English evaluation under a shared task definition.

\section{\medrect{} Benchmark}

\subsection{Task Definition}

Following \medec{}, we decompose medical error detection and correction into three progressive subtasks (Figure~\ref{fig:task_overview}): (1) \textbf{Error Detection}, binary classification to determine whether a clinical text contains an error, (2) \textbf{Error Sentence Extraction}, identifying the specific sentence with the error, and (3) \textbf{Error Correction}, generating a corrected version of the erroneous sentence. This decomposition enables fine-grained evaluation of model capabilities and helps identify weaknesses in the three-stage evaluation. The latter two subtasks apply only to clinical texts that contain an error. Figure~\ref{fig:examples_compact} illustrates examples across different error types in both \medrect{}-ja and \medrect{}-en.

\begin{figure}[!htb]
\centering
\begin{tikzpicture}[
    examplebox/.style={
        rectangle, draw, rounded corners=2pt,
        text width=0.95\columnwidth,
        align=left, font=\scriptsize,
        inner sep=4pt,
        fill=gray!5
    },
    errorhighlight/.style={
        rectangle, draw=none, rounded corners=1pt,
        inner sep=1pt
    },
    typelabel/.style={
        rectangle, draw, rounded corners=1.5pt,
        font=\tiny\bfseries,
        inner sep=1.5pt,
        anchor=north east
    }
]

\node[examplebox] (ex1) at (0,0) {
    \textbf{Example 1 (\medrect{}-ja):}\\[0.1em]
    A 20-day-old male infant was brought to the hospital by his mother with chief complaints of poor feeding and fever. [...] Physical examination: Poor complexion, bulging anterior fontanelle, and irritability were observed. Cerebrospinal fluid: Cell count 4,200/mm³, protein 80 mg/dL, glucose 5 mg/dL. {\setlength{\fboxsep}{1pt}\colorbox{orange!20}{The laboratory findings strongly suggested meningitis caused}}\\
    {\setlength{\fboxsep}{1pt}\colorbox{orange!20}{by \textbf{\color{red!70!black}\textit{Pseudomonas aeruginosa}}.}}\\[0.1em]
    \textbf{Reference correction:} The laboratory findings strongly suggested meningitis caused by {\color{ForestGreen}\textbf{GBS (\textit{Streptococcus agalactiae})}}.
};
\node[typelabel, fill=orange!15] at (ex1.north east) {Diagnosis};

\node[examplebox, below=0.25cm of ex1] (ex2) {
    \textbf{Example 2 (\medrect{}-ja):}\\[0.1em]
    An 82-year-old woman is hospitalized in a palliative care unit for pancreatic cancer with liver metastases. For the past week, she has had a progressive loss of appetite and decreased food intake. [...] {\setlength{\fboxsep}{1pt}\colorbox{purple!20}{The nutrition support team proposed the \textbf{\color{red!70!black}placement of a gastrostomy}}}\\
    {\setlength{\fboxsep}{1pt}\colorbox{purple!20}{\textbf{\color{red!70!black}tube} for nutritional supplementation.}} The policy was to explain the plan to the attending physician and obtain consent.\\[0.1em]
    \textbf{Reference correction:} The nutrition support team decided to prioritize consideration of {\color{ForestGreen}\textbf{non-invasive nutritional management}} methods.
};
\node[typelabel, fill=purple!15] at (ex2.north east) {Procedures/intervention};

\node[examplebox, below=0.25cm of ex2] (ex3) {
    \textbf{Example 3 (\medrect{}-en):}\\[0.1em]
    A 73-year-old man presents with 'weird blisters' on his right hand, appeared 2 weeks ago. [...] {\setlength{\fboxsep}{1pt}\colorbox{teal!20}{\textbf{\color{red!70!black}Doxycycline} is prescribed after physical exam.}} On physical exam: multiple bullae with red, papular lesions on right hand progressing to forearm. Right axillary lymph nodes are swollen and tender.\\[0.1em]
    \textbf{Reference correction:} {\color{ForestGreen}\textbf{Itraconazole}} is prescribed after physical exam.
};
\node[typelabel, fill=teal!15] at (ex3.north east) {Medication selection};

\end{tikzpicture}
\caption{Examples from the \medrect{} dataset showing different error types. Examples 1 and 2 show \medrect{}-ja samples (translated to English for readability), while Example 3 shows a native \medrect{}-en sample derived from \medec{} \citep{abacha2025medecbenchmarkmedicalerror}. Each example highlights the erroneous sentence (colored background), the key error in \textbf{\color{red!70!black}red}, and the reference correction in \textbf{\color{ForestGreen}green}.}
\label{fig:examples_compact}
\end{figure}

\subsection{Dataset Construction}
\label{subsec:data_construction_pipeline}

Both \medrect{}-ja and \medrect{}-en consist of structured clinical vignettes (chief complaints, examination findings, laboratory results, and clinical decision points) derived from medical licensing examinations rather than free-form clinical notes or EHR records. \medrect{}-ja is constructed directly from JMLE rather than translated English data, grounding evaluation in Japanese-language vignettes and the institutional context of Japanese medical education and licensure. The two subsets share a task definition but are neither matched nor translated counterparts, so they support language-specific evaluation rather than direct estimates of language or cultural effects. Both subsets underwent automated screening but not systematic clinician adjudication.

For \medrect{}-ja, we extracted 287 clinical case questions from JMLE (2024--2025) and applied four stages (see Figure~\ref{fig:data_pipeline} in Appendix for the complete pipeline): (1) automatic transformation with error injection (287$\rightarrow$2,792 samples), (2) consensus-based difficulty filtering ($\rightarrow$1,057), (3) deduplication ($\rightarrow$720), and (4) LLM-as-a-Judge quality screening ($\rightarrow$663).

During development of the Japanese pipeline, the first author, a medical student in clinical clerkship, manually inspected sample quality throughout the process, including the final retained set, and used these inspections to iteratively adjust parameters throughout the pipeline, including prompts, filtering thresholds, and screening criteria.

For \medrect{}-en, we applied the same LLM-as-a-Judge screening framework with source-adapted criteria to the \medec{} MS test subset (597 samples)~\citep{ben-abacha-etal-2024-overview,abacha2025medecbenchmarkmedicalerror}, producing 458 samples.

The automated Japanese construction pipeline is designed for reuse with other exam sources. In this work, the English subset is instead curated from existing manually annotated \medec{} data. Complete procedures are detailed in Appendix~\ref{subsec:pipeline_details}.

\subsection{Dataset Statistics}

\begin{table}[!htb]
\centering
\small
\caption{Dataset statistics for \medrect{}-ja and \medrect{}-en}
\label{tab:dataset_stats_compact}
\setlength{\tabcolsep}{3pt}
\begin{tabular}{lcc}
\toprule
& \medrect{}-ja & \medrect{}-en \\
\midrule
Total samples & 663 & 458 \\
With errors & 367 (55.4\%) & 243 (53.1\%) \\
Without errors & 296 (44.6\%) & 215 (46.9\%) \\
\midrule
\multicolumn{3}{l}{\textit{Error Type Distribution}} \\
Diagnosis & 77 (21.0\%) & 98 (40.3\%) \\
Monitoring/management & 79 (21.5\%) & 17 (7.0\%) \\
Physical findings & 72 (19.6\%) & 2 (0.8\%) \\
Procedures/intervention & 40 (10.9\%) & 38 (15.6\%) \\
Medication selection & 30 (8.2\%) & 70 (28.8\%) \\
Test interpretation & 37 (10.1\%) & 12 (4.9\%) \\
History taking & 22 (6.0\%) & 1 (0.4\%) \\
Medication dosage & 8 (2.2\%) & 3 (1.2\%) \\
Other & 2 (0.5\%) & 2 (0.8\%) \\
\bottomrule
\end{tabular}
\end{table}

\medrect{}-ja contains 663 samples with 367 (55.4\%) errors and 296 (44.6\%) correct texts, while \medrect{}-en comprises 458 samples with 243 (53.1\%) errors and 215 (46.9\%) correct texts. The two subsets have similar error-to-correct ratios of approximately 55:45.

Error type distributions reflect different clinical contexts and source methodologies. \medrect{}-ja shows balanced distributions across diagnosis (21.0\%), monitoring/management (21.5\%), and physical findings (19.6\%), reflecting the breadth of clinical scenarios covered by the JMLE. \medrect{}-en is dominated by diagnosis errors (40.3\%) and medication selection (28.8\%), reflecting the underlying MedQA source patterns.

For analysis, we use the eight error types specified in the Japanese synthesis prompt. Gemini 2.5 Pro~\citep{gemini25_2025} reassigned the original \medec{} labels to this taxonomy. Two residual Japanese labels and two unmapped English outputs are reported as \textit{Other}.

\paragraph{Privacy and Content Safety}
All source data from JMLE are publicly available standardized test materials that do not contain personally identifiable information (PII). For \medrect{}-en, we use the \medec{} MS test subset~\citep{ben-abacha-etal-2024-overview}, which was constructed from MedQA medical licensing exam questions~\citep{jin2020diseasedoespatienthave} and similarly contains no PII.

\section{Experimental Setup}

\subsection{Evaluated Models}
\label{subsec:evaluated_models}
We evaluated \nummodels{} contemporary LLMs across \numconfigs{} configurations. \textbf{Reasoning models} (GPT-5~\citep{openai_gpt5_systemcard_pdf_2025}, o3~\citep{openai_o3_2025}, Claude Sonnet 4~\citep{anthropic_claude4_systemcard_2025}, gpt-oss-120b and gpt-oss-20b~\citep{openai_gptoss_modelcard_arxiv_2025}, and Qwen3-32B in thinking mode~\citep{qwen3_2025}) employ explicit step-by-step reasoning during inference. \textbf{Non-reasoning models} (GPT-4.1~\citep{openai_gpt41_2025}, DeepSeek-V3-0324~\citep{deepseek_v3_2024}, and Qwen3-32B in non-thinking mode) do not. \textbf{Medical-domain models} (HuatuoGPT-o1-72B~\citep{chen2025huatuogpto1}, MedResearcher-R1-32B~\citep{li2025medresearcherr1}, MedGemma 27B~\citep{medgemma2025}) are specifically trained or fine-tuned for medical tasks. Qwen3-32B supports two inference modes, thinking and non-thinking (labeled think and no-think in the tables), within the same architecture, enabling a within-model comparison of reasoning modes. We list it in both categories accordingly. OpenAI's reasoning models (GPT-5, o3, gpt-oss) support configurable reasoning effort parameters within computational token limits. We used the API default medium setting for GPT-5 and o3, while evaluating gpt-oss across all three levels (high/medium/low). Claude Sonnet 4 uses a thinking parameter that enables extended thinking, which we enabled for evaluation.
\subsection{Evaluation Metrics}

We report error detection F1, error sentence extraction accuracy, and three reference-based correction similarity metrics: ROUGE-1~\citep{rouge}, BERTScore~\citep{bertscore}, and BLEURT~\citep{bleurt}. We treat ERROR as the positive class for error detection F1. An always-ERROR predictor yields F1 scores of 0.713 and 0.693 on \medrect{}-ja and \medrect{}-en, respectively. We report the arithmetic average of the three correction metrics as a compact descriptive summary while retaining all three component scores. ROUGE-1 uses MeCab~\citep{kudo-etal-2004-applying} for Japanese and whitespace tokenization for English. BERTScore F1 uses microsoft/mdeberta-v3-base (no baseline rescaling) so that the same multilingual encoder scores both languages, and BLEURT uses BLEURT-20.

Error sentence extraction accuracy and reference-based correction similarity scores are computed on samples with a ground-truth error. The similarity scores are averaged over all such samples, with missed detections (the model outputs \texttt{CORRECT} for an error-containing sample) assigned 0. This avoids inflating scores for low-recall models that would otherwise be evaluated only on the errors they happened to flag. The Appendix \medec{} comparison (Table~\ref{tab:appendix_medec_compact}) instead follows the original MEDIQA-CORR 2024 evaluation protocol~\citep{ben-abacha-etal-2024-overview} for comparability with reported physician baselines.

For comparisons between Qwen3-32B in thinking and non-thinking modes and between Qwen3-32B in thinking mode with and without LoRA, we estimate source-cluster uncertainty with 20,000 paired nonparametric bootstrap resamples (seed 20260715) and report 95\% percentile confidence intervals (CIs). Predictions are aligned by sample ID. For \medrect{}-ja, samples are clustered by source JMLE question. For \medrect{}-en, each \medec{} Text ID is a separate source record. Error detection F1 is resampled over all source clusters, and error sentence extraction accuracy over source clusters containing at least one ground-truth error. These intervals quantify source-record variation for fixed predictions, not run-to-run generation variability.

\subsection{Evaluation Prompts}

Models were evaluated using carefully designed zero-shot prompts that instructed medical experts to identify and correct a clinical error. The evaluation prompt is shown below:

\begin{tcolorbox}[title={Evaluation Prompt (0\_shot\_en)}, boxrule=0.5pt, colback=gray!5, top=2pt, bottom=2pt, left=3pt, right=3pt, breakable]
\scriptsize
\setlength{\baselineskip}{9pt}
\texttt{You are a medical expert reviewing clinical text for accuracy. The text contains either no error or exactly one medical error.\\
\\
Identify and correct any medical error related to treatment, diagnosis, management, or causation.\\
\\
Output Format:\\
- If no error: `CORRECT`\\
- If error found: `sentence\_number: corrected\_sentence`\\
\\
CRITICAL: Output ONLY the result. Do NOT include explanations, analysis, or additional text.\\
\\
\{sentences\}}
\end{tcolorbox}

For Japanese evaluation, we directly translated the prompt (0\_shot\_ja), preserving the same task specification and output format.

\subsection{Fine-tuning Configuration}

Fine-tuning was performed using LoRA with rank=64, $\alpha$=128 on Qwen3-32B as the underlying model. We employed a learning rate of \texttt{1e-4} for effective task adaptation.

Qwen3-32B was fine-tuned on combined Japanese (5,538 samples) and English (2,439 samples) training data with reasoning traces generated by DeepSeek-R1-0528 (see Appendix~\ref{sec:training_data} for details).

\begin{table*}[t!]
\centering
\footnotesize
\setlength{\tabcolsep}{2.8pt}
\caption{Bilingual performance for \numconfigs{} configurations of \nummodels{} LLMs on \medrect{}-ja and \medrect{}-en.
Parentheses indicate reasoning effort levels for gpt-oss (high, medium, or low)
and reasoning modes for Qwen3-32B (think or no-think).
}
\label{tab:performance_bilingual}
\resizebox{\textwidth}{!}{%
\begin{tabular}{lcccccccccccc}
\toprule
& \multicolumn{6}{c}{\medrect{}-ja} & \multicolumn{6}{c}{\medrect{}-en} \\
\cmidrule(lr){2-7} \cmidrule(lr){8-13}
Model
& {\scriptsize Error Detection} & {\scriptsize Sentence Extraction} & \multicolumn{4}{c}{\scriptsize \shortstack{Reference-based\\correction similarity}}
& {\scriptsize Error Detection} & {\scriptsize Sentence Extraction} & \multicolumn{4}{c}{\scriptsize \shortstack{Reference-based\\correction similarity}} \\
\cmidrule(lr){4-7} \cmidrule(lr){10-13}
& {\scriptsize F1} & {\scriptsize Acc. (\%)} & {\scriptsize ROUGE-1} & {\scriptsize BERTScore} & {\scriptsize BLEURT} & {\scriptsize Avg.}
& {\scriptsize F1} & {\scriptsize Acc. (\%)} & {\scriptsize ROUGE-1} & {\scriptsize BERTScore} & {\scriptsize BLEURT} & {\scriptsize Avg.} \\
\midrule
\multicolumn{13}{l}{\textit{Reasoning models}} \\
GPT-5               & .758 & \textbf{83.7} & .502 & .712 & .518 & .577   & .818 & \textbf{96.3} & \textbf{.689} & \textbf{.848} & \textbf{.697} & \textbf{.745} \\
o3                   & .764 & 71.4 & .454 & .636 & .458 & .516   & \textbf{.852} & 87.7 & .631 & .781 & .644 & .685 \\
Claude Sonnet 4      & \textbf{.795} & 82.3 & \textbf{.559} & \textbf{.755} & \textbf{.547} & \textbf{.620}   & .784 & 84.0 & .642 & .800 & .643 & .695 \\
gpt-oss-120b (high)  & .731 & 79.6 & .476 & .733 & .509 & .572   & .759 & 92.6 & .648 & .841 & .670 & .719 \\
gpt-oss-120b (medium)   & .721 & 77.4 & .424 & .690 & .469 & .528   & .777 & 88.1 & .593 & .796 & .608 & .666 \\
gpt-oss-120b (low)   & .704 & 68.9 & .390 & .667 & .434 & .497   & .775 & 79.4 & .552 & .751 & .568 & .624 \\
gpt-oss-20b (high)   & .729 & 71.4 & .429 & .690 & .461 & .527   & .757 & 86.0 & .569 & .782 & .593 & .648 \\
gpt-oss-20b (medium)    & .718 & 64.3 & .371 & .651 & .413 & .478   & .762 & 87.2 & .527 & .754 & .558 & .613 \\
gpt-oss-20b (low)    & .678 & 46.9 & .281 & .584 & .335 & .400   & .723 & 71.2 & .394 & .633 & .434 & .487 \\
Qwen3-32B + LoRA (think) & .743 & 81.5 & .494 & .715 & .479 & .563   & .728 & 90.9 & .670 & .825 & .648 & .714 \\
Qwen3-32B (think)    & .723 & 72.5 & .368 & .646 & .431 & .482   & .740 & 83.5 & .495 & .748 & .549 & .597 \\
\midrule
\multicolumn{13}{l}{\textit{Non-reasoning models}} \\
GPT-4.1              & .658 & 52.6 & .304 & .429 & .317 & .350   & .789 & 72.8 & .527 & .646 & .521 & .565 \\
DeepSeek-V3-0324     & .688 & 42.2 & .324 & .624 & .361 & .437   & .684 & 42.0 & .387 & .667 & .437 & .497 \\
Qwen3-32B (no-think) & .637 & 48.0 & .249 & .532 & .300 & .360   & .704 & 73.3 & .442 & .706 & .488 & .545 \\
\midrule
\multicolumn{13}{l}{\textit{Medical-domain models}} \\
HuatuoGPT-o1-72B     & .702 & 62.1 & .356 & .547 & .378 & .427   & .721 & 68.3 & .504 & .674 & .511 & .563 \\
MedResearcher-R1-32B & .558 & 35.4 & .219 & .364 & .235 & .272   & .663 & 61.7 & .435 & .577 & .417 & .476 \\
MedGemma 27B         & .603 & 39.2 & .273 & .481 & .282 & .346   & .658 & 64.6 & .494 & .642 & .466 & .534 \\
\bottomrule
\end{tabular}
}
\end{table*}

\section{Results}

\subsection{Performance across Reasoning Settings}
\label{subsec:reasoning_results}

Table~\ref{tab:performance_bilingual} presents bilingual results for \numconfigs{} configurations of \nummodels{} models on \medrect{}-ja and \medrect{}-en. All numbers in this subsection refer to \medrect{}-ja unless otherwise noted. Claude Sonnet 4 achieves both the highest average reference-based correction similarity (0.620) and the highest error detection F1 (0.795). The next-highest averages are GPT-5 (0.577), gpt-oss-120b at high reasoning effort (0.572), and Qwen3-32B with LoRA in thinking mode (0.563). Error detection F1 ranges from 0.558 (MedResearcher-R1-32B) to 0.795 (Claude Sonnet 4), while sentence extraction accuracy ranges from 35.4\% to 83.7\%.

Two within-model comparisons show higher scores under higher-reasoning settings. First, average reference-based correction similarity falls as reasoning effort decreases for both gpt-oss model sizes. For gpt-oss-120b, it is 0.572 at high effort, 0.528 at medium effort, and 0.497 at low effort. Second, comparing Qwen3-32B in thinking and non-thinking modes holds model weights and architecture fixed, although the decoding settings also differ. On \medrect{}-ja, the thinking and non-thinking modes yield 0.723 vs.\ 0.637 error detection F1 and 72.5\% vs.\ 48.0\% sentence extraction accuracy. The paired differences are +0.086 (95\% CI [0.051, 0.124]) and +24.5 percentage points (95\% CI [17.5, 31.7]). On \medrect{}-en, the corresponding differences are +0.037 (95\% CI [0.008, 0.066]) and +10.3 percentage points (95\% CI [4.9, 15.6]) (Appendix Table~\ref{tab:paired_bootstrap}).

Several leading general-purpose reasoning models score above all three evaluated medical-domain models on error detection and sentence extraction. HuatuoGPT-o1-72B~\citep{chen2025huatuogpto1}, the strongest medical-domain model, achieves 62.1\% sentence extraction accuracy, below the 72.5\% of the smaller Qwen3-32B in thinking mode. MedGemma 27B~\citep{medgemma2025} achieves 39.2\%. These cross-family comparisons do not isolate reasoning from model size or training data. Together with the within-model results, they show that medical-domain specialization alone does not ensure stronger performance on this benchmark.

\begin{table*}[t!]
\centering
\scriptsize
\setlength{\tabcolsep}{2.8pt}
\caption{Error sentence extraction accuracy (\%) by error type on \medrect{}-ja and \medrect{}-en.
For gpt-oss, only medium reasoning effort is shown. The four samples categorized as \textit{Other} are excluded.
Results based on fewer than 10 samples should be interpreted cautiously.
Best values are bolded only for columns with at least 10 samples.
}
\label{tab:error_types_bilingual}
\begin{tabular}{lcccccccccccccccc}
\toprule
{\footnotesize} & \multicolumn{2}{c}{\footnotesize Diagnosis} & \multicolumn{2}{c}{\footnotesize Monitoring/} & \multicolumn{2}{c}{\footnotesize Physical} & \multicolumn{2}{c}{\footnotesize Procedures/} & \multicolumn{2}{c}{\footnotesize Test} & \multicolumn{2}{c}{\footnotesize Medication} & \multicolumn{2}{c}{\footnotesize History} & \multicolumn{2}{c}{\footnotesize Medication} \\
& \multicolumn{2}{c}{} & \multicolumn{2}{c}{\footnotesize management} & \multicolumn{2}{c}{\footnotesize findings} & \multicolumn{2}{c}{\footnotesize intervention} & \multicolumn{2}{c}{\footnotesize interpretation} & \multicolumn{2}{c}{\footnotesize selection} & \multicolumn{2}{c}{\footnotesize taking} & \multicolumn{2}{c}{\footnotesize dosage} \\
\cmidrule(lr){2-3} \cmidrule(lr){4-5} \cmidrule(lr){6-7} \cmidrule(lr){8-9} \cmidrule(lr){10-11} \cmidrule(lr){12-13} \cmidrule(lr){14-15} \cmidrule(lr){16-17}
& {\footnotesize ja} & {\footnotesize en} & {\footnotesize ja} & {\footnotesize en} & {\footnotesize ja} & {\footnotesize en} & {\footnotesize ja} & {\footnotesize en} & {\footnotesize ja} & {\footnotesize en} & {\footnotesize ja} & {\footnotesize en} & {\footnotesize ja} & {\footnotesize en} & {\footnotesize ja} & {\footnotesize en} \\[2pt]
\multicolumn{1}{r}{{\footnotesize\textit{\char`\#samples}}} & {\footnotesize\textit{77}} & {\footnotesize\textit{98}} & {\footnotesize\textit{79}} & {\footnotesize\textit{17}} & {\footnotesize\textit{72}} & {\footnotesize\textit{2}} & {\footnotesize\textit{40}} & {\footnotesize\textit{38}} & {\footnotesize\textit{37}} & {\footnotesize\textit{12}} & {\footnotesize\textit{30}} & {\footnotesize\textit{70}} & {\footnotesize\textit{22}} & {\footnotesize\textit{1}} & {\footnotesize\textit{8}} & {\footnotesize\textit{3}} \\
\midrule
\multicolumn{17}{l}{\textit{Reasoning models}} \\
GPT-5                   & 90.9 & \textbf{99.0} & \textbf{89.9} & \textbf{100.0} & 72.2 & 100.0 & \textbf{97.5} & \textbf{89.5} & 70.3 & \textbf{100.0} & \textbf{100.0} & 94.3 & 45.5 & 100.0 & 100.0 & 100.0 \\
o3                      & 80.5 & 94.9 & 73.4 & 58.8 & 56.9 & 100.0 & 90.0 & 76.3 & 56.8 & 91.7 & \textbf{100.0} & 88.6 & 27.3 & 100.0 & 100.0 & 100.0 \\
Claude Sonnet 4         & 92.2 & 89.8 & 84.8 & 64.7 & 75.0 & 100.0 & 92.5 & 68.4 & 67.6 & 75.0 & 83.3 & 88.6 & 63.6 & 100.0 & 100.0 & 100.0 \\
gpt-oss-120b (medium) & 81.8 & 91.8 & 69.6 & 82.4 & 72.2 & 50.0 & 87.5 & 84.2 & \textbf{81.1} & 75.0 & \textbf{100.0} & 88.6 & 45.5 & 100.0 & 100.0 & 100.0 \\
gpt-oss-20b (medium)  & 76.6 & 84.7 & 59.5 & 76.5 & 50.0 & 100.0 & 77.5 & 78.9 & 62.2 & 91.7 & 83.3 & \textbf{95.7} & 27.3 & 100.0 & 87.5 & 100.0 \\
Qwen3-32B + LoRA (think)  & \textbf{94.8} & 92.9 & 75.9 & \textbf{100.0} & \textbf{83.3} & 100.0 & 87.5 & 86.8 & 70.3 & 91.7 & 80.0 & 90.0 & \textbf{81.8} & 0.0 & 12.5 & 66.7 \\
Qwen3-32B (think)       & 72.7 & 85.7 & 74.7 & 76.5 & 58.3 & 100.0 & 82.5 & 73.7 & 78.4 & 83.3 & 93.3 & 87.1 & 59.1 & 100.0 & 62.5 & 66.7 \\
\midrule
\multicolumn{17}{l}{\textit{Non-reasoning models}} \\
GPT-4.1            & 54.5 & 76.5 & 64.6 & 64.7 & 51.4 & 100.0 & 60.0 & 63.2 & 37.8 & 83.3 & 70.0 & 71.4 & 4.5 & 100.0 & 37.5 & 66.7 \\
DeepSeek-V3-0324        & 45.5 & 48.0 & 57.0 & 29.4 & 44.4 & 50.0 & 40.0 & 31.6 & 18.9 & 16.7 & 30.0 & 47.1 & 40.9 & 0.0 & 25.0 & 66.7 \\
Qwen3-32B (no-think)    & 58.4 & 81.6 & 45.6 & 64.7 & 45.8 & 50.0 & 47.5 & 57.9 & 51.4 & 75.0 & 43.3 & 72.9 & 40.9 & 100.0 & 12.5 & 66.7 \\
\midrule
\multicolumn{17}{l}{\textit{Medical-domain models}} \\
HuatuoGPT-o1-72B       & 74.0 & 80.6 & 64.6 & 64.7 & 41.7 & 50.0 & 72.5 & 52.6 & 54.1 & 58.3 & 83.3 & 62.9 & 40.9 & 0.0 & 87.5 & 66.7 \\
MedResearcher-R1-32B    & 51.9 & 63.3 & 31.6 & 47.1 & 26.4 & 50.0 & 40.0 & 47.4 & 27.0 & 50.0 & 56.7 & 74.3 & 13.6 & 100.0 & 0.0 & 33.3 \\
MedGemma 27B            & 44.2 & 70.4 & 36.7 & 58.8 & 34.7 & 0.0 & 30.0 & 55.3 & 43.2 & 83.3 & 46.7 & 64.3 & 36.4 & 0.0 & 62.5 & 33.3 \\
\bottomrule
\end{tabular}
\end{table*}

\subsection{Bilingual Comparison and Fine-tuning}
\label{subsec:bilingual_finetuning}

Because \medrect{}-ja and \medrect{}-en are constructed from different source materials (JMLE vs.\ MedQA-derived \medec{}) and have different error type distributions (Table~\ref{tab:dataset_stats_compact}), absolute scores across the two language subsets are not directly comparable. We therefore focus on within-language rankings and within-model differences. Most models score lower on \medrect{}-ja, but the magnitude varies substantially. For sentence extraction, GPT-4.1 scores 52.6\% on \medrect{}-ja and 72.8\% on \medrect{}-en, while error detection F1 is similar for Claude Sonnet 4 (0.795 vs.\ 0.784). Within each language, the evaluated medical-domain models also trail the leading reasoning models on sentence extraction and average reference-based correction similarity.

To control for differences in error type distribution between \medrect{}-ja and \medrect{}-en (Table~\ref{tab:dataset_stats_compact}), we also report macro-averaged sentence extraction accuracy with equal weighting across the eight named error types in Table~\ref{tab:error_types_bilingual}, excluding \textit{Other}. For models with uneven per-type performance, macro-averaging reveals notable shifts: Qwen3-32B with LoRA in thinking mode drops from 81.5\% (micro) to 73.3\% (macro) on \medrect{}-ja, primarily due to 12.5\% accuracy on medication dosage ($n{=}8$), while GPT-5 remains stable (83.7\% micro vs.\ 83.3\% macro). Under macro-averaging, most models still score lower on \medrect{}-ja than on \medrect{}-en (e.g., GPT-5: 83.3\% vs.\ 97.8\%, and Qwen3-32B with LoRA in thinking mode: 73.3\% vs.\ 78.5\%), indicating that the direction of within-model differences is not merely an artifact of the error type distribution. Macro-averaged \medrect{}-en scores should be interpreted cautiously given small sample sizes in several error categories.

LoRA fine-tuning yields higher sentence extraction accuracy and higher point estimates on all three reference-based correction similarity metrics in both languages. Our fine-tuned model is publicly available (Appendix~\ref{subsec:released_artifacts}). On \medrect{}-ja, Qwen3-32B with LoRA in thinking mode achieves 0.563 average reference-based correction similarity compared with 0.482 without LoRA in the same mode (16.8\% relative improvement). Sentence extraction accuracy increases by 9.0 percentage points (95\% CI [2.8, 15.0]). The error detection F1 difference is +0.020 (95\% CI [-0.011, 0.051]).

On \medrect{}-en, average reference-based correction similarity increases from 0.597 to 0.714 (19.6\% relative improvement), and sentence extraction accuracy increases by 7.4 percentage points (95\% CI [2.5, 12.3]) to 90.9\%. The error detection F1 difference is -0.012 (95\% CI [-0.040, 0.016]). Thus, the sentence extraction differences exclude zero in both languages, while the error detection differences do not.

Under the original \medec{} evaluation protocol, the fine-tuned model achieves 90.6\% sentence extraction accuracy compared with 76.7\% and 64.6\% for two physicians, and 0.714 average reference-based correction similarity compared with 0.491 and 0.678 (Appendix Table~\ref{tab:appendix_medec_compact}). Its error detection accuracy is lower (62.0\% vs.\ 81.3\% and 68.9\%) due to more false positives. These results should be interpreted cautiously given the two-person human baseline and the restricted task setting.

\begin{table*}[!t]
\centering
\small
\caption{Model outputs on three selected \medrect{} samples}
\label{tab:qualitative_examples_compact}
\resizebox{0.98\textwidth}{!}{
\scriptsize
\begin{tabular}{|p{1.5cm}|p{3.5cm}|p{3.3cm}|p{4.5cm}|}
\hline
\textbf{Sample} & \textbf{Sample 1: 119B36\_a\_deepseek-r1-0528 (ja)} & \textbf{Sample 2: 118E37\_c\_qwen3-235b-a22b-thinking-2507 (ja)} & \textbf{Sample 3: ms-test-120 (en)} \\
\hline
\textbf{Error Type} & \textit{Procedures/intervention} & \textit{History taking} & \textit{(No error)} \\
\hline
\textbf{Clinical Text} & 82-year-old female admitted to palliative care ward due to pancreatic cancer with liver metastasis. Progressive appetite loss and decreased food intake for 1 week [...]. Physical findings: decreased skin turgor, poor oral hygiene and dryness [...]. Blood findings: [...] Hb 9.2g/dL [...]. Biochemistry: [...] albumin 2.8g/dL [...]. Nutrition support team proposed \textbf{\color{red!70!black}gastrostomy} for nutritional support. Plan to explain to attending physician and obtain consent. & 80-year-old female diagnosed with lung cancer and multiple bone metastases 1 month ago. She chose not to receive active treatment and receives home visits [...]. She mainly stays in bed during the day. Living alone since husband's death at age 65, with distant contact from son [...]. She told the visiting physician "I want to end it all." The physician asked \textbf{\color{red!70!black}"What is the most distressing physical symptom you have right now?"} The patient did not respond and remained lying in bed. & 45-year-old woman presents complaining of weakness that has progressively worsened over the past 2 weeks. She states that she has a hard time lifting both her arms but that they function normally [...]. On examination, patient has 2/5 muscle strength on shoulder shrug and arm abduction bilaterally, but all other neurological exam findings are normal. You notice some skin changes [...]. She states that she has had a rash around her eyes as well as on her lower face, going down to her neck and chest. She notes that the rashes started around the same time as the weakness began. Patient is diagnosed with dermatomyositis after labs are drawn and a complete blood count and basic metabolic panel are normal. \\
\hline

\parbox[t]{1.5cm}{\textbf{Reference \\ Output}} & Nutrition support team should prioritize consideration of {\color{ForestGreen}\textbf{non-invasive nutrition management}} methods. & The physician responded {\color{ForestGreen}\textbf{"You're feeling distressed. Could you tell me more about your feelings?"}} & CORRECT \\
\hline

\textbf{GPT-5} & Nutrition support team judged that gastrostomy is inappropriate given the terminal cancer with limited prognosis, and proposed \textbf{non-invasive nutritional support} prioritizing symptom palliation with oral intake strategies, oral care, and supplementary fluids as needed. & The physician first assessed for \textbf{suicidal ideation and urgency} (specific plans, means, intent), ensured safety, and then comprehensively assessed physical, psychological, social, and spiritual distress. & Patient is diagnosed with dermatomyositis after labs show \textbf{elevated muscle enzymes} (e.g., creatine kinase), while the complete blood count and basic metabolic panel may be normal. \\
\hline

\parbox[t]{1.5cm}{\textbf{Claude \\ Sonnet 4}} & Nutrition support team proposed \textbf{oral intake support} for symptom palliation. & The physician asked \textbf{"Could you tell me more about those feelings?"} & Patient is diagnosed with dermatomyositis after labs are drawn including \textbf{elevated muscle enzymes and positive autoantibodies}, while complete blood count and basic metabolic panel are normal. \\
\hline

\parbox[t]{1.5cm}{\textbf{Qwen3-32B \\ +~LoRA \\ (think)}} & Nutrition support team proposed \textbf{coordination with home care}, prioritizing continuation of oral intake above all else. & The physician responded \textbf{"You have been carrying this alone and must be very tired."} & \textbf{CORRECT} \\
\hline

\textbf{Qwen3-32B (think)} & Nutrition support team is considering introduction of \textbf{enteral nutrition (gastrostomy placement)} as a nutritional supplement method based on palliative care goals and the patient's own decision-making. & The physician \textbf{failed to properly evaluate} the patient's statement and focused solely on physical symptoms, potentially missing psychosocial factors. & Patient is diagnosed with dermatomyositis after labs show \textbf{elevated creatine kinase, erythrocyte sedimentation rate}, and characteristic skin findings consistent with heliotrope rash and shawl sign. \\
\hline
\end{tabular}
}

\begin{tablenotes}
\tiny
\item \textbf{\color{red!70!black}Red} in Clinical Text highlights the key error. {\color{ForestGreen}\textbf{Green}} in Reference Output highlights the correction where applicable.
\item \medrect{}-ja samples (119B36\_a\_deepseek-r1-0528 and 118E37\_c\_qwen3-235b-a22b-thinking-2507) are translated to English for readability.
\item For Sample 2, the reference reproduces the JMLE-keyed best initial reply. It does not represent a complete suicide-risk assessment. Outputs are shown without clinical correctness ratings.
\end{tablenotes}
\end{table*}

\subsection{Performance by Error Type}

Performance breakdown across different medical error categories reveals substantial variation in task difficulty and model behavior patterns (Table~\ref{tab:error_types_bilingual}). All numbers in this subsection refer to \medrect{}-ja sentence extraction accuracy unless otherwise noted.

\paragraph{Error Type Difficulty}
\textit{History taking} has the lowest average accuracy across models on \medrect{}-ja, with values ranging from 4.5\% to 81.8\%. This suggests that contextual understanding of patient interactions is difficult under the evaluated settings. \textit{Physical findings} and \textit{test interpretation} also have relatively low scores. In contrast, most reasoning models achieve 80\% to 100\% on \textit{medication selection}. \textit{Medication dosage} varies from 0.0\% to 100.0\%, but this category contains only eight Japanese samples and three English samples.

\paragraph{Reasoning and Fine-tuning Comparisons}
Comparing Qwen3-32B in thinking and non-thinking modes shows large gaps in \textit{medication selection} (93.3\% vs.\ 43.3\%, +50.0 percentage points) and \textit{procedures/intervention} (82.5\% vs.\ 47.5\%, +35.0 percentage points) under the evaluated settings. LoRA fine-tuning in thinking mode produces the largest gains in \textit{history taking} (+22.7 percentage points) and \textit{physical findings} (+25.0 percentage points). The observed drop in \textit{medication dosage} (62.5\% to 12.5\%) is based on eight samples, so it cannot establish overfitting or a systematic fine-tuning effect. The gpt-oss model comparison also shows that parameter count alone does not order category-level performance in these settings.

\subsection{Qualitative Analysis}\label{sec:qualitative}

Table~\ref{tab:qualitative_examples_compact} presents three cases and their model outputs, covering palliative care, communication with a distressed patient, and false positive correction. These examples describe individual outputs and are not a systematic clinical evaluation.

In Sample 1, most models identify gastrostomy placement as inappropriate for a terminally ill patient with limited prognosis. The fine-tuned model and proprietary systems propose non-invasive alternatives that prioritize comfort care.

In Sample 2, GPT-5 proposes direct assessment of suicidal ideation and safety, Claude Sonnet 4 invites further disclosure, and the LoRA fine-tuned model offers an empathic reflection. The JMLE-derived reference specifies the keyed best initial reply rather than a complete management plan. This case illustrates that reference-based similarity does not fully capture clinical adequacy.

In Sample 3, GPT-5, Claude Sonnet 4, and Qwen3-32B in thinking mode without LoRA incorrectly flag an accurate text and propose unnecessary additions. The LoRA fine-tuned model returns no correction. This example illustrates how a false positive could create unnecessary review work in a clinical workflow.

\section{Discussion}

On \medrect{}-ja, sentence extraction accuracy varies from 35.4\% to 83.7\% across settings. Higher-reasoning settings generally outperform lower-reasoning settings within the gpt-oss and Qwen3 families, although low-effort configurations can overlap the non-reasoning range. Leading general-purpose reasoning models also outperform the medical-domain models, indicating that domain specialization alone does not ensure strong performance.

For Qwen3-32B, sentence extraction accuracy is 24.5 percentage points higher in thinking mode on \medrect{}-ja and 10.3 higher on \medrect{}-en. LoRA raises average reference-based correction similarity by 0.081 (16.8\% relative) and 0.117 (19.6\% relative), respectively. Contrasting within-model patterns across \medrect{}-ja and \medrect{}-en highlight the value of evaluation across language, source, and medical-education contexts. Because the subsets are not item-aligned, the patterns indicate context-specific behavior rather than isolated language or cultural effects. \medrect{}-ja provides a non-translated benchmark grounded in Japanese medical education and licensure.

The examples in Section~\ref{sec:qualitative} illustrate clinically relevant behaviors beyond reference-based correction similarity, but do not establish general clinical capability.

\section{Conclusion}

We introduce \medrect{}, an open bilingual benchmark for medical error detection and correction. Across \nummodels{} LLMs, higher-reasoning settings are associated with stronger error detection and sentence extraction, while leading general-purpose reasoning models outperform the medical-domain alternatives. LoRA fine-tuning improves sentence extraction accuracy and all three reference-based correction similarity point estimates in both subsets.

\section{Limitations}
\label{sec:limitations}

\paragraph{Dataset Validation}
\medrect{} was constructed without systematic clinician adjudication. During development, manual quality checks by the first author, a medical student in clinical clerkship, included the final retained set and informed iterative refinement of the Japanese pipeline (Section~\ref{subsec:data_construction_pipeline}). Automated screening applied the resulting criteria (Appendix~\ref{subsec:pipeline_details}) to remove problems such as internal contradictions in clinical values and ambiguous error labels (Appendix~\ref{subsec:quality_screening_results}). This medically informed oversight does not establish clinical correctness for every sample or replace independent clinician review. The pipeline uses constrained synthetic transformations grounded in exam materials developed by medical expert committees. Systematic clinician validation remains future work.

\paragraph{Dataset and Scope}
The dataset size is constrained by the availability of suitable JMLE questions (287 of 800 across two examination years yielded 663 samples after synthesis and filtering), and our synthetic error generation may not fully cover the diversity of real clinical errors. \medrect{} also focuses on text-based scenarios with single-sentence errors. Real-world errors may span multiple sentences, depend on document-level context (e.g., contradictory lab values across sections), or involve multimodal documents.

\paragraph{Reference-Based Evaluation}
The correction metrics measure similarity to a single reference correction. They do not directly assess clinical correctness, sufficiency, or safety. As Sample 2 in Table~\ref{tab:qualitative_examples_compact} illustrates, a clinically defensible response may add safety assessment beyond the keyed reference and receive lower similarity.

\paragraph{Model-Specific Biases}
The construction pipeline uses specific models at multiple steps. The two synthesis models (DeepSeek-R1-0528 and Qwen3-235B-A22B-Thinking-2507) and the quality-screening model (Gemini 2.5 Pro) are excluded from evaluation. The consensus-based difficulty filtering in Step 2 uses 11 configurations across 8 models rather than single-model judgments. GPT-4.1 and Qwen3-32B variants are used in both consensus filtering and evaluation. Appendix~\ref{subsec:filter_sensitivity} reports a conservative sensitivity check, but the unavailable item-level historical votes prevent exact reconstruction of a fully disjoint filtered benchmark. Results for reused models should therefore be interpreted with this limitation. Some released English training traces retain meta-references to provided answer information despite prompt instructions to avoid them, so these traces should not be interpreted as independently elicited rationales.

\paragraph{Cross-Language Comparability}
\medrect{}-ja and \medrect{}-en are constructed from different source materials (JMLE vs.\ MedQA-derived \medec{}) and have different error type distributions (Table~\ref{tab:dataset_stats_compact}). Absolute scores on the two language subsets therefore reflect differences in source material and item difficulty as well as language, and should not be read as a direct measure of cross-lingual transfer. Throughout the paper we use \medrect{}-ja and \medrect{}-en scores for within-language rankings and within-model comparisons (e.g., pre- vs.\ post-fine-tuning), and we avoid strong claims about the absolute magnitude of cross-language performance differences.

\paragraph{Potential Risks}
Performance varies substantially across error types, with particularly low sentence extraction accuracy for medication dosage errors (as low as 0.0\% and 12.5\% in Table~\ref{tab:error_types_bilingual}), highlighting risks in high-stakes clinical domains. Most models also score lower on \medrect{}-ja than on \medrect{}-en (Table~\ref{tab:performance_bilingual}). Without attributing this gap to language alone, the pattern motivates separate evaluation and development for non-English medical contexts. \medrect{} is designed for research evaluation, not direct clinical deployment. Any real-world application should maintain human-in-the-loop validation of all automated predictions.

\section*{Use of Generative AI}
AI coding assistants (e.g., Claude Code) were used by the authors to support the implementation of data processing pipelines, evaluation scripts, and analysis tooling. All AI-generated code was reviewed and tested by the authors before inclusion.

\section*{Acknowledgments}
We are grateful to the Preferred Networks cluster team for the computational infrastructure support.

\bibliography{references}

\appendix

\section{\medrect{} Dataset Construction}

\subsection{Pipeline Details}
\label{subsec:pipeline_details}

This section provides comprehensive technical details for the data construction pipeline described in Section~\ref{subsec:data_construction_pipeline}. Figure~\ref{fig:data_pipeline} shows the complete pipeline for both \medrect{}-ja and \medrect{}-en.

\begin{figure*}[!htb]
\centering
\resizebox{\textwidth}{!}{%
\begin{tikzpicture}[
    node distance=0.9cm and 0.35cm,
    box/.style={
        rectangle, draw, rounded corners=1.5pt,
        text width=1.6cm,
        align=center, minimum height=1.1cm, font=\tiny
    },
    stagebox/.style={box, fill=blue!15},
    sourcebox/.style={box, fill=purple!15},
    finalbox/.style={box, fill=green!15},
    arrow/.style={->, >=Stealth, semithick, black}
]

\node[sourcebox] (jmle) {
    \textbf{JMLE}\\
    2024 \& 2025\\
    \textit{287 questions}
};
\node[stagebox, right=of jmle] (synthesis) {
    \textbf{Synthesis}\\
    \textbf{2,792}\\
    ~
};
\node[stagebox, right=of synthesis] (filtering) {
    \textbf{Consensus filtering}\\
    \textbf{1,057}\\
    \textcolor{red}{(-1,735)}
};
\node[stagebox, right=of filtering] (deduplication) {
    \textbf{Deduplication}\\
    \textbf{720}\\
    \textcolor{red}{(-337)}
};
\node[stagebox, right=of deduplication] (quality_ja) {
    \textbf{Quality screening}\\
    \textbf{663}\\
    \textcolor{red}{(-57)}
};
\node[finalbox, right=of quality_ja] (finalja) {
    \textbf{\medrect{}-ja}\\
    \textbf{663} samples
};

\node[sourcebox, below=of jmle] (medqa) {
    \textbf{MedQA}\\
    Medical Board\\
    Examinations
};
\node[stagebox, below=of deduplication] (medec) {
    \textbf{\medec{}}\\
    MS test subset\\
    \textbf{597} samples
};
\node[stagebox, below=of quality_ja] (quality_en) {
    \textbf{Quality screening}\\
    \textbf{458}\\
    \textcolor{red}{(-139)}
};
\node[finalbox, below=of finalja] (finalen) {
    \textbf{\medrect{}-en}\\
    \textbf{458} samples
};

\node[font=\small\bfseries, above=0.4cm of jmle] {\medrect{}-ja Construction};
\node[font=\small\bfseries, above=0.4cm of medqa] {\medrect{}-en Construction};

\draw[arrow] (jmle) -- (synthesis);
\draw[arrow] (synthesis) -- (filtering);
\draw[arrow] (filtering) -- (deduplication);
\draw[arrow] (deduplication) -- (quality_ja);
\draw[arrow] (quality_ja) -- (finalja);

\draw[arrow] (medqa) -- (medec);
\draw[arrow] (medec) -- (quality_en);
\draw[arrow] (quality_en) -- (finalen);

\node[font=\scriptsize, text width=3.5cm, align=center] at ($(medqa.south)!0.5!(medec.north)$) {
    Medical annotators\\
    \textbf{heuristically} create clinical texts
};

\end{tikzpicture}
}
\caption{Data construction pipeline for \medrect{}. \medrect{}-ja (top) transforms JMLE questions through automated synthesis, consensus-based difficulty filtering, model deduplication, and LLM-as-a-Judge quality screening. \medrect{}-en (bottom) applies the same screening framework with source-adapted criteria to the existing \medec{} MS test subset. Red numbers indicate samples removed at each step.}
\label{fig:data_pipeline}
\end{figure*}

\subsubsection{Automated Synthesis}

We extracted 287 clinical case questions from JMLE (2024--2025), focusing on questions that described patient scenarios while excluding image-based questions, calculation problems, and questions with underlined text that would complicate reformatting. For each JMLE question, we generated clinical texts by incorporating each answer choice into the original clinical scenario, creating CORRECT samples (from correct choices) and ERROR samples (from wrong choices).

We used two reasoning LLMs for synthesis: DeepSeek-R1-0528~\citep{deepseek_r1_2025} and Qwen3-235B-A22B-Thinking-2507~\citep{qwen3_2025}.

The synthesis prompt requested one of eight error types: history taking, physical findings, test interpretation, diagnosis, monitoring/management, medication selection, medication dosage, and procedures/intervention. Each model was asked for five records per question (2,870 requested). Parsing the available complete and partial responses yielded 2,792 candidate samples.

\subsubsection{Consensus-Based Difficulty Filtering}

We evaluated each synthesized sample on error detection and error sentence extraction using eight validation models: Gemini 2.5 Pro~\citep{gemini25_2025}, GPT-4.1~\citep{openai_gpt41_2025}, PLaMo 2.0 Prime~\citep{pfn2025plamo2technicalreport}, Qwen3-8B, Qwen3-14B, Qwen3-32B, Qwen3-30B-A3B-Thinking-2507~\citep{qwen3_2025}, and QwQ-32B~\citep{qwq32b,qwen2.5}. Qwen3-8B, Qwen3-14B, and Qwen3-32B were evaluated in both thinking and non-thinking modes, yielding 11 configurations.

We counted a complete response as correct when it returned CORRECT for a no-error sample or identified the gold error sentence for an ERROR sample. Across all samples, we retained cases with complete-response accuracy between $1/11$ and $7/11$ and a gap of at most $3/11$ between error detection accuracy and complete-response accuracy. This excluded cases that were uniformly easy or difficult under the validation configurations and cases where models detected an error without reliably locating it.

This process filtered the dataset from 2,792 to 1,057 samples with intermediate difficulty under the validation configurations.

\subsubsection{Model Deduplication}

Since both synthesis models (DeepSeek-R1-0528 and Qwen3-235B-A22B-Thinking-2507) processed identical JMLE source questions, our pipeline generated duplicate samples from the same source (question, answer choice) pairs. For each duplicate pair, we retained one sample and alternated the retained source model across pairs to balance representation. This reduced the dataset from 1,057 to 720 samples.

\subsubsection{LLM-as-a-Judge Quality Screening}

We employed LLM-as-a-Judge (Gemini 2.5 Pro) to perform binary classification on quality dimensions. For \medrect{}-ja, we assessed five dimensions:
\begin{itemize}
    \item \textit{ambiguous\_error}: Medical statements with unclear correctness
    \item \textit{extra\_elements}: Addition of information not in original problem/choices
    \item \textit{multiple\_errors}: Multiple error locations in ERROR data
    \item \textit{numerical\_error}: Numerical errors difficult to correct from context
    \item \textit{synthesis\_consistency\_error}: Wrong choice used but medically correct content
\end{itemize}

For \medrect{}-en, because samples originated from existing \medec{} data without original JMLE questions, we adapted the criteria by replacing \textit{extra\_elements} and \textit{synthesis\_consistency\_error} with two analogous dimensions:
\begin{itemize}
    \item \textit{unrealistic\_scenario}: Clinically implausible scenarios
    \item \textit{inconsistent\_context}: Internal inconsistencies in clinical context
\end{itemize}

Any sample scoring 1 (problematic) on any dimension was excluded from the final dataset. This screening produced 663 samples for \medrect{}-ja (92.1\% retention) and 458 samples for \medrect{}-en (76.7\% retention).

\subsection{Data Synthesis Prompt}
\label{subsec:data_synthesis}

An abridged English translation of the synthesis prompt used to transform JMLE questions into clinical texts is shown below. The synthesis example is omitted for brevity.

\begin{tcolorbox}[title={Data Synthesis Prompt (English Translation)}, boxrule=0.5pt, colback=gray!5, top=2pt, bottom=2pt, left=3pt, right=3pt, breakable]
\scriptsize
\setlength{\baselineskip}{9pt}
\texttt{Convert the following Japanese medical licensing examination question into clinical cases in MEDEC (Medical Error Detection and Correction) format.\\
\\
\# Instructions\\
- For each answer choice, synthesize one clinical record incorporating that choice into the problem text, creating 5 records total.\\
- If the choice is correct, synthesize a correct record; if wrong, synthesize a record containing an error.\\
- Clinical records should always be written as numbered markdown lists, with error-containing records having exactly one sentence with a clinical error.\\
- These records will be used for MEDEC format benchmarks. Do not indicate where errors are located in error-containing records.\\
- Include all numerical values and findings from the original problem without summarization or omission.\\
- Do not add original medical interpretations not present in the original problem or choices.\\
\\
\# Original Medical Licensing Examination Problem\\
Problem: \{question\}\\
Choices: \{choices\_text\}\\
Correct choices: \{correct\_choices\_list\}\\
Wrong choices: \{wrong\_choices\_list\}\\
\\
\# Synthesis Format\\
\\
The following shows examples of correct records synthesized from correct choices and error records from wrong choices:\\
\\
\#\#\# Choice \{correct\_choices\_list\}[0] Record (CORRECT sample)\\
1. ...\\
2. ...\\
...\\
N. ...\\
\\
\#\#\# Choice \{wrong\_choices\_list\}[0] Record (ERROR sample)\\
1. ...\\
2. ...\\
...\\
N. ...\\
Error Type: [Select from: history taking, physical findings, test interpretation, diagnosis, medication selection, medication dosage, procedures/intervention, monitoring/management]\\
Error Sentence Number: [Number of sentence containing the medical error]\\
Error Sentence: [The sentence with wrong medical content]\\
Corrected Sentence: [The medically accurate version of the sentence]\\
\\
(Synthesis example omitted for brevity.)}
\end{tcolorbox}

\subsection{Quality Screening Prompt}
\label{subsec:quality_screening}

An abridged English translation of the no-demonstration quality-screening prompt used with Gemini 2.5 Pro is shown below. We also used a four-demonstration variant with the same quality dimensions.

\begin{tcolorbox}[title={Quality Screening Prompt (No Demonstrations, Abridged English Translation)}, boxrule=0.5pt, colback=gray!5, top=2pt, bottom=2pt, left=3pt, right=3pt, breakable]
\scriptsize
\setlength{\baselineskip}{9pt}
\texttt{\#\# MEDEC Benchmark Quality Assessment\\
\\
\#\#\# Original Medical Examination Problem\\
Question: \{original\_question\}\\
Choices: \{choices\_text\}\\
Correct answer: \{correct\_answer\}\\
**Used choice: \{used\_choice\_text\}**\\
\\
\#\#\# Generated MEDEC Format Text\\
\{sentences\}\\
\\
\#\#\# Error Information\\
\{error\_info\}\\
\\
\#\#\# Assessment Task\\
Please evaluate the **quality as a benchmark problem** for the above generated text.\\
\\
Rate the following aspects as 1 (problematic) or 0 (acceptable):\\
\\
1. **ambiguous\_error**: Medical statements with unclear correctness\\
2. **extra\_elements**: Addition of information not in original problem/choices\\
3. **multiple\_errors**: Multiple error locations in ERROR data\\
4. **numerical\_error**: Numerical errors difficult to correct from context\\
5. **synthesis\_consistency\_error**: Wrong choice used but medically correct content\\
\\
JSON response: \{"ambiguous\_error": 0, "extra\_elements": 0, "multiple\_errors": 0, "numerical\_error": 0, "synthesis\_consistency\_error": 0, "explanation": "Brief assessment"\}}
\end{tcolorbox}

\vspace{0.5em}
\noindent \textbf{Note:} For \medrect{}-en dataset construction, the above criteria were adapted to account for differences in source material characteristics. Specifically, \textit{extra\_elements} and \textit{synthesis\_consistency\_error} were replaced with \textit{unrealistic\_scenario} and \textit{inconsistent\_context} to better suit the existing exam-derived vignettes in the \medec{} dataset.

\subsection{Quality Screening Results}
\label{subsec:quality_screening_results}

We applied both the no-demonstration and four-demonstration prompting configurations to each sample. Any sample that scored 1 (problematic) on any quality dimension in either configuration was excluded from the final dataset, yielding different retention rates for the two language subsets (Table~\ref{tab:quality_screening_results_compact}).

\begin{table}[!htb]
\centering
\small
\caption{Quality screening results and exclusion reasons. Reasons are non-mutually exclusive, so their counts can exceed the number of excluded samples.}
\label{tab:quality_screening_results_compact}
\setlength{\tabcolsep}{3pt}
\begin{tabular}{lcc}
\toprule
& \medrect{}-ja & \medrect{}-en \\
\midrule
Original samples & 720 & 597 \\
Retained samples & 663 (92.1\%) & 458 (76.7\%) \\
\midrule
\multicolumn{3}{l}{\textit{Exclusion Reasons}} \\
Ambiguous error & 3 & 98 \\
Extra elements & 9 & -- \\
Multiple errors & 21 & 24 \\
Numerical error & 1 & 7 \\
Synthesis consistency & 27 & -- \\
Unrealistic scenario & -- & 28 \\
Inconsistent context & -- & 58 \\
\bottomrule
\end{tabular}
\end{table}

For \medrect{}-en, the screening removed 139 of 597 samples (23.3\%), primarily due to ambiguous errors (98 samples). This filtering targets objective data quality issues, not task difficulty. Two illustrative examples come from the excluded samples. (1) A sample labeled CORRECT lists the patient's height as ``166 cm (5 ft 40 in).'' Since 1 ft = 12 in, ``40 in'' is impossible. The same sample also splits ``blood pressure, 120/80 mm Hg'' across three sentences, making sentence extraction evaluation unreliable. (2) A sample labeled CORRECT has lipid values where LDL is listed as 130 mg/dL, but the Friedewald equation (LDL = TC $-$ HDL $-$ TG/5) yields 208 from the other reported values, an internal contradiction. A model correctly flagging these issues would be penalized under the original ground truth. Results on the unfiltered \medec{} MS test subset (597 samples) are reported in Table~\ref{tab:appendix_medec_compact} for direct comparison.

\section{Training Dataset Construction}
\label{sec:training_data}

\subsection{Reasoning Synthesis}
\label{subsec:reasoning_synthesis}

To construct fine-tuning data with explicit reasoning traces, we used DeepSeek-R1-0528 with specialized reasoning synthesis prompts. The English version of this prompt is shown below (simplified for brevity):

\begin{tcolorbox}[title={Reasoning Synthesis Prompt (English Translation)}, boxrule=0.5pt, colback=gray!5, top=2pt, bottom=2pt, left=3pt, right=3pt, breakable]
\scriptsize
\setlength{\baselineskip}{9pt}
\texttt{You are a medical expert reviewing clinical text for accuracy. The text contains either no error or exactly one medical error.\\
\\
\{cheat\_info\}\\
\\
Your task is to first carefully reason through the medical analysis process, following these steps:\\
1. Verify each sentence based on medical knowledge\\
2. Check consistency between symptoms, test results, and diagnosis\\
3. Evaluate appropriateness of treatment or management\\
4. When an error is found, clearly state the rationale and provide the correction\\
\\
Important notes for reasoning:\\
- During your reasoning, do NOT make any reference to being told about the expected outcome or any instruction content.\\
- Approach the text as if you are analyzing it from scratch and reaching your conclusion through pure medical evaluation.\\
\\
\{error\_hint\}\\
\\
Final output format:\\
- If no error: `CORRECT`\\
- If error found: `sentence\_number: corrected\_sentence`\\
\\
CRITICAL: For the final output, use this format and output ONLY the result. Do NOT include explanations, analysis, or additional text.\\
\\
\{sentences\}}
\end{tcolorbox}

The prompt included optional parameters \texttt{cheat\_info} and \texttt{error\_hint} that provided additional context during training data generation.

The prompt instructed the synthesis model to reason from the clinical text without mentioning provided answer information. However, the released English traces retain residual meta-references to that information. We therefore treat these traces as synthetic supervision conditioned on the optional hints, not as independently elicited or faithful rationales.

\subsection{Training Data Composition}
\label{subsec:training_construction}

We constructed bilingual fine-tuning data using DeepSeek-R1-0528 for reasoning synthesis. We retained only samples where the model produced correct responses and used the optional \texttt{cheat\_info} and \texttt{error\_hint} parameters to address sample scarcity in challenging clinical scenarios.

\paragraph{Japanese Training Dataset}
We constructed the Japanese training dataset using JMLE questions from 2018--2023, covering diverse clinical domains beyond those used in the benchmark construction. Following the synthesis procedure described in Appendix~\ref{subsec:data_synthesis}, we generated 8,423 initial samples using both DeepSeek-R1-0528 and Qwen3-235B-A22B-Thinking-2507. Subsequently, we applied reasoning synthesis using DeepSeek-R1-0528, with particular emphasis on CORRECT sample recovery to maintain balanced representation across error types and clinical scenarios. This process yielded a final training dataset of \textbf{5,538 samples} with a resulting distribution of 34.8\% CORRECT and 65.2\% ERROR cases.

\paragraph{English Training Dataset}
For English training data, we used the \medec{} MS training and validation subsets, containing 2,763 expert-annotated clinical texts. These samples underwent reasoning synthesis using DeepSeek-R1-0528 with the same prompts used for the Japanese dataset. The resulting English training dataset comprised \textbf{2,439 samples}, with 49.0\% CORRECT and 51.0\% ERROR cases.

\section{Additional Results}

Table~\ref{tab:paired_bootstrap} reports the full paired source-cluster bootstrap results for the within-model comparisons discussed in Sections~\ref{subsec:reasoning_results} and~\ref{subsec:bilingual_finetuning}. The Japanese benchmark contains 663 samples from 281 retained JMLE source questions, including 188 questions represented by multiple samples. The English benchmark contains 458 singleton source records.

\paragraph{Difficulty-Filter Sensitivity}\label{subsec:filter_sensitivity}
Three of the 11 validation configurations used for difficulty filtering (GPT-4.1 and Qwen3-32B in thinking and non-thinking modes) also appear in the final evaluation. The individual historical votes were not retained, so we cannot exactly refilter all 2,792 candidates after removing these configurations. We instead used the retained aggregate counts to identify a conservative core guaranteed to remain within proportionally scaled eight-configuration thresholds for any allocation of the three overlapping votes. Specifically, the retained count must contain four or five complete-response votes and a detection--complete-response gap of at most two. This core contains 121 samples (37 ERROR) from 102 JMLE questions. On this core, the difference between Qwen3-32B in thinking and non-thinking modes is +0.156 error detection F1 (95\% source-cluster CI [0.004, 0.307]) and +27.0 percentage points in sentence extraction accuracy (95\% CI [5.3, 48.7]). The difference between Qwen3-32B with and without LoRA in thinking mode is +0.065 error detection F1 (95\% CI [-0.043, 0.181]) and +27.0 percentage points in sentence extraction accuracy (95\% CI [5.1, 50.0]). The intervals are wide because the core contains only 37 error samples. This one-sided check cannot recover rejected candidates that might become eligible after refiltering, so it reduces but does not eliminate the selection concern.

\begin{table*}[t!]
\centering
\footnotesize
\setlength{\tabcolsep}{3pt}
\caption{Paired source-cluster bootstrap differences for focused within-model configuration changes. Each difference is the right-hand configuration minus the left-hand configuration. Brackets show 95\% percentile confidence intervals from 20,000 resamples. Sentence extraction differences are percentage points.}
\label{tab:paired_bootstrap}
\resizebox{\textwidth}{!}{%
\begin{tabular}{llcc}
\toprule
Configuration change & Benchmark & Error detection F1 difference & Sentence extraction accuracy difference \\
\midrule
Qwen3-32B: (no-think) $\rightarrow$ (think) & \medrect{}-ja & +0.086 [0.051, 0.124] & +24.5 [17.5, 31.7] \\
Qwen3-32B: (no-think) $\rightarrow$ (think) & \medrect{}-en & +0.037 [0.008, 0.066] & +10.3 [4.9, 15.6] \\
Qwen3-32B (think): without LoRA $\rightarrow$ with LoRA & \medrect{}-ja & +0.020 [-0.011, 0.051] & +9.0 [2.8, 15.0] \\
Qwen3-32B (think): without LoRA $\rightarrow$ with LoRA & \medrect{}-en & -0.012 [-0.040, 0.016] & +7.4 [2.5, 12.3] \\
\bottomrule
\end{tabular}
}
\end{table*}

\begin{table*}[t!]
\centering
\begin{threeparttable}
\footnotesize
\setlength{\tabcolsep}{5.5pt}
\caption{Performance on the unfiltered \medec{} MS test subset.
Parentheses give gpt-oss reasoning effort or Qwen3-32B reasoning mode.
}
\label{tab:appendix_medec_compact}
\begin{tabular}{lccccccc}
\toprule
\multirow{2}{*}{Model} & \multicolumn{2}{c}{Error Detection} & Sentence Extraction & \multicolumn{4}{c}{\shortstack{Reference-based correction similarity\tnote{$\dagger$}}} \\
\cmidrule(lr){2-3} \cmidrule(lr){4-4} \cmidrule(lr){5-8}
& F1 & Acc. (\%) & Acc. (\%) & ROUGE-1 & BERTScore & BLEURT & Avg. \\
\midrule
\multicolumn{8}{l}{\textit{MEDEC Paper Results}} \\
Medical Doctor \#1 & - & \textbf{81.3} & 76.7 & 0.420 & 0.513 & 0.539 & 0.491 \\
Medical Doctor \#2 & - & 68.9 & 64.6 & 0.685 & 0.698 & 0.650 & 0.678 \\
\midrule
\multicolumn{8}{l}{\textit{Reasoning models}} \\
GPT-5 & 0.780 & 71.7 & \textbf{90.7} & 0.655 & 0.672 & 0.671 & 0.666 \\
o3 & \textbf{0.783} & 75.0 & 80.7 & 0.658 & 0.680 & 0.677 & 0.672 \\
Claude Sonnet 4 & 0.737 & 67.2 & 75.2 & 0.640 & 0.667 & 0.653 & 0.653 \\
gpt-oss-120b (high) & 0.733 & 62.5 & 87.1 & 0.606 & 0.626 & 0.633 & 0.621 \\
gpt-oss-120b (medium) & 0.742 & 66.2 & 82.0 & 0.582 & 0.605 & 0.606 & 0.598 \\
gpt-oss-120b (low) & 0.740 & 67.5 & 74.0 & 0.566 & 0.594 & 0.592 & 0.584 \\
gpt-oss-20b (high) & 0.726 & 63.3 & 78.8 & 0.554 & 0.581 & 0.590 & 0.575 \\
gpt-oss-20b (medium) & 0.736 & 65.5 & 83.3 & 0.540 & 0.573 & 0.579 & 0.564 \\
gpt-oss-20b (low) & 0.694 & 63.1 & 67.8 & 0.458 & 0.495 & 0.515 & 0.489 \\
Qwen3-32B + LoRA (think) & 0.723 & 62.0 & 90.6 & \textbf{0.711} & \textbf{0.748} & \textbf{0.684} & \textbf{0.714} \\
Qwen3-32B (think) & 0.711 & 60.5 & 77.5 & 0.480 & 0.509 & 0.546 & 0.512 \\
\midrule
\multicolumn{8}{l}{\textit{Non-reasoning models}} \\
GPT-4.1 & 0.726 & 72.9 & 65.6 & 0.683 & 0.697 & 0.681 & 0.687 \\
DeepSeek-V3-0324 & 0.671 & 54.6 & 38.6 & 0.399 & 0.428 & 0.470 & 0.432 \\
Qwen3-32B (no-think) & 0.688 & 57.6 & 69.8 & 0.461 & 0.486 & 0.517 & 0.488 \\
\bottomrule
\end{tabular}
\begin{tablenotes}
\footnotesize
\item[$\dagger$] For direct comparison with the published physician baselines,
this table follows the original MEDIQA-CORR 2024 protocol: BERTScore uses
microsoft/deberta-xlarge-mnli with English baseline rescaling, and the three
reference-based correction similarity metrics are averaged over the
(ground-truth = error, prediction = error) intersection. Table~\ref{tab:performance_bilingual}
instead uses microsoft/mdeberta-v3-base without rescaling and assigns 0 to
missed detections over all samples with a ground-truth error. Values are not directly
comparable across the two tables.
\end{tablenotes}
\end{threeparttable}
\end{table*}

Table~\ref{tab:appendix_medec_compact} reports performance on the unfiltered \medec{} MS test subset (597 samples) before quality screening (see Appendix~\ref{subsec:quality_screening_results}). To enable direct comparison with the physician baselines from the original \medec{} paper~\citep{abacha2025medecbenchmarkmedicalerror} and the MEDIQA-CORR 2024 shared task~\citep{ben-abacha-etal-2024-overview}, both the BERTScore configuration and the reference-based correction similarity aggregation in this table follow the original protocol rather than the setup used in Table~\ref{tab:performance_bilingual}. Reference-based correction similarity scores are therefore not directly comparable across the two tables (see table note).

\section{Computational Resources and Licensing}
\label{sec:resources_licensing}

\subsection{Computational Resources}
\label{subsec:compute_resources}

LoRA fine-tuning on Qwen3-32B and inference for the open-weight models (gpt-oss and Qwen3-32B) and medical-domain models in Section~\ref{subsec:evaluated_models} were performed on single NVIDIA A100 80GB GPU nodes. Proprietary models (GPT-5, GPT-4.1, o3, and Claude Sonnet 4) and DeepSeek-V3-0324 were accessed via the OpenRouter API and were not run on local hardware. Parameter counts follow the original releases cited in Section~\ref{subsec:evaluated_models}.

\subsection{Source Data Licensing}
\label{subsec:source_data_licensing}

\medrect{}-ja is derived from the Japanese Medical Licensing Examinations (JMLE 2024--2025), published by the Ministry of Health, Labour and Welfare of Japan under the Public Data License 1.0 (PDL-1.0), which is compatible with CC BY 4.0. \medrect{}-en is derived from the MS test subset of \medec{}~\citep{abacha2025medecbenchmarkmedicalerror}, released by its authors at \medecrepo{} under CC BY 4.0. We do not use the UW subset, which requires a separate data usage agreement. LLM-as-a-Judge quality screening applied to both language subsets is documented in Appendix~\ref{subsec:quality_screening_results}.

\subsection{Released Artifacts}
\label{subsec:released_artifacts}

Our datasets are released under CC BY 4.0 (compatible with both source licenses), and the accompanying code under the MIT License; both are available at \medrectrepo{}. Complete evaluation outputs are archived on Zenodo at \medrectzenodo{}. Our LoRA fine-tuned model is released under Apache 2.0 (inheriting from Qwen3-32B) and available at \medrectmodel{}.

\end{document}